\documentclass[11pt,a4paper]{article}
\usepackage[hyperref]{naaclhlt2019}
\usepackage{times}
\usepackage{latexsym}
\usepackage{graphicx}
\usepackage{eurosym}

\usepackage{kantlipsum}
\usepackage{tabularx}
\graphicspath{ {images/} }
\usepackage{hyperref}
\usepackage{url}
\usepackage{amsmath}
\usepackage{hhline}
\usepackage[utf8]{inputenc}

\usepackage[T1]{fontenc}
\usepackage{graphicx}
\usepackage{eurosym}
\usepackage{kantlipsum}
\usepackage{listings}
\usepackage{stfloats}
\usepackage{csquotes}
\usepackage{array}
\usepackage{booktabs}
\usepackage[super]{nth}
\usepackage[inline,shortlabels]{enumitem}
\usepackage{multirow}
\usepackage{enumitem}
\usepackage{makecell}

\usepackage{svg}
\usepackage{float}
\usepackage{ amssymb }

\usepackage{marginnote}
\usepackage{mdframed}
\usepackage{xprintlen}
\usepackage{todonotes}
\usepackage{url}

\usepackage{pgfplots}

\usepackage{multicol}
\usepackage{subfig}
\pgfplotsset{compat=1.9}

\aclfinalcopy 

\title{An Embarrassingly Simple Approach for Transfer Learning from Pretrained Language Models}

\author{ Alexandra Chronopoulou$^{1}$, Christos Baziotis$^{1}$, Alexandros Potamianos$^{1,2,3}$ \\
	$^1$School of ECE, National Technical University of Athens, Athens, Greece\\
	$^2$ Signal Analysis and Interpretation Laboratory (SAIL), USC, Los Angeles, USA\\
	$^3$ Behavioral Signal Technologies, Los Angeles, USA\\
        	{\tt  el12068@central.ntua.gr, cbaziotis@mail.ntua.gr} \\
	{\tt  potam@central.ntua.gr}\\
		}
\date{}

\begin{document}
\maketitle

\begin{abstract}

A growing number of state-of-the-art transfer learning methods employ language models pretrained on large generic corpora. In this paper we present a conceptually simple and effective transfer learning approach that addresses the problem of catastrophic forgetting. Specifically, we combine the task-specific optimization function with an auxiliary language model objective, which is adjusted during the training process. This preserves language regularities captured by language models, while enabling sufficient adaptation for solving the target task. Our method does not require pretraining or finetuning separate components of the network and we train our models end-to-end in a single step. We present results on a variety of challenging affective and text classification tasks, surpassing well established transfer learning methods with greater level of complexity.
\end{abstract}
\section{Introduction}

Pretrained word representations captured by Language Models (LMs) have recently become popular in Natural Language Processing (NLP). Pretrained LMs encode contextual information and high-level features of language, modeling syntax and semantics, producing state-of-the-art results across a wide range of tasks, such as named entity recognition \cite{peters2017semi}, machine translation \cite{ramachandran2016unsupervised} and text classification \cite{howard2018universal}.

However, in cases where contextual embeddings from language models are used as additional features (e.g. ELMo \cite{peters2018deep}), results come at a high computational cost and require task-specific architectures. At the same time, approaches that rely on fine-tuning a LM to the task at hand (e.g. ULMFiT \cite{howard2018universal}) 
depend on pretraining the model on an extensive vocabulary and on
employing a sophisticated slanted triangular learning rate scheme  to adapt the parameters of the LM to the target dataset.

We propose a simple and effective transfer learning approach, that leverages LM contextual representations and does not require any elaborate scheduling schemes during training. 
We initially train a LM on a Twitter corpus and then transfer its weights. We add a task-specific recurrent layer and a classification layer. The transferred model is trained end-to-end using an auxiliary LM loss, which allows us to explicitly control the weighting of the pretrained part of the model and ensure that the distilled knowledge it encodes is preserved. 

Our contributions are summarized as follows: 1) We show that transfer learning from language models can achieve competitive results, while also being intuitively simple and computationally effective. 2) We address the problem of catastrophic forgetting, by adding an auxiliary LM objective and using an unfreezing method. 3) Our results show that our approach is competitive with more sophisticated transfer learning methods.
We make our code widely available.~\footnote{\url{www.github.com/alexandra-chron/siatl}}

\section{Related Work}

Unsupervised pretraining has played a key role in deep neural networks, building on the premise that representations learned for one task can be useful for another task. In NLP, pretrained word vectors \cite{mikolov2013, pennington2014} are widely used, improving performance in various downstream tasks, such as part-of-speech tagging \cite{collobert2011natural} and question answering \cite{xiong2016dynamic}. These pretrained word vectors serve as initialization of the embedding layer and remain frozen during training, while our pretrained language model also initializes the hidden layers of the model and is fine-tuned to each classification task.

Aiming to learn from unlabeled data, \citet{dai2015semi} use unsupervised objectives such as sequence autoencoding and language modeling for as pretraining methods. The pretrained model is then fine-tuned to the target task. However, the fine-tuning procedure of the language model to the target task does not include an auxiliary objective. \citet{ramachandran2016unsupervised} also pretrain encoder-decoder pairs using language models and fine-tune them to a specific task, using an auxiliary language modeling objective to prevent catastrophic forgetting. This approach, nevertheless, is only evaluated on machine translation tasks; moreover, the seq2seq \cite{sutskever2014sequence} and language modeling losses are weighted equally throughout training. By contrast, we propose a weighted sum of losses, where the language modeling contribution gradually decreases.
ELMo embeddings \cite{peters2018deep} are obtained from language models and improve the results in a variety of tasks as additional contextual representations. However, ELMo embeddings rely on character-level models, whereas our approach uses a word-level LM. They are, furthermore, concatenated to pretrained word vectors and remain fixed during training. We instead propose a fine-tuning procedure, aiming to adjust a generic architecture to different end tasks.

Moreover, BERT \cite{devlin2018bert} pretrains language models and fine-tunes them on the target task. An auxiliary task (next sentence prediction) is used to enhance the representations of the LM. BERT fine-tunes masked bi-directional LMs. Nevertheless, we are limited to a uni-directional model. Training BERT requires vast computational resources, while our model only requires 1 GPU. We note that our approach is not orthogonal to BERT and could be used to improve it, by adding an auxiliary LM objective and weighing its contribution.

Towards the same direction, ULMFiT \cite{howard2018universal} shows impressive results on a variety of tasks by employing pretrained LMs. The proposed pipeline requires three distinct steps, that include (1) pretraining the LM, (2) fine-tuning it on a target dataset with an elaborate scheduling procedure and (3) transferring it to a classification model. Our proposed model is closely related to ULMFiT. However, ULMFiT trains a LM and fine-tunes it to the target dataset, before transferring it to a classification model. While fine-tuning the LM to the target dataset, the metric (e.g. accuracy) that we intend to optimize cannot be observed. We propose adopting a multi-task learning perspective, via the addition of an auxiliary LM loss to the transferred model, to control the loss of the pretrained and the new task simultaneously. The intuition is that we should avoid catastrophic forgetting, but at the same time allow the LM to distill the knowledge of the prior data distribution and keep the most useful features.

Multi-Task Learning (MTL) via hard parameter sharing \cite{caruna1993multitask} in neural networks has proven to be effective in many NLP problems \cite{collobert2008unified}. More recently, alternative approaches have been suggested that only share parameters across lower layers \cite{sogaard2016deep}. By introducing part-of-speech tags at the lower levels of the network, the proposed model achieves competitive results on chunking and CCG super tagging. Our auxiliary language model objective follows this line of thought and intends to boost the performance of the higher classification layer.

\section{Our Model}\label{sec:four}

We introduce \textbf{SiATL}, which stands for \textbf{Si}ngle-step \textbf{A}uxiliary loss \textbf{T}ransfer \textbf{L}earning. In our proposed approach, we first train a LM. We then transfer its weights and add a task-specific recurrent layer to the final classifier. We also employ an auxiliary LM loss to avoid catastrophic forgetting.

\vspace{4pt}

\noindent\textbf{LM Pretraining.}
We train a word-level language model, which consists of an embedding LSTM layer \cite{hochreiter1997long}, 2 hidden LSTM layers and a linear layer. We want to minimize the negative log-likelihood of the LM:
\begin{align}
L(\hat{p}) &= -\frac{1}{N}\sum_{n=1}^{N}\sum_{t=1}^{T^n} \text{log}\hat{p}(x_t^n|x_1^n,...,x_{t-1}^n)
\end{align}
where $\hat{p}(x_t^n|x_1^n,...,x_{t-1}^n)$ is the distribution of the $t^{th}$ word in the $n^{th}$ sentence given the $t - 1$ words preceding it and $N$ is total number of sentences. 
\vspace{5pt}

\noindent\textbf{Transfer \& auxiliary loss.}
\label{transferlm}
We  transfer the weights of the pretrained model and add one LSTM with a self-attention mechanism \cite{lin2017structured,bahdanau2014neural}. 
\begin{figure}[t]
	\captionsetup{farskip=0pt} 
	\centering
	\includegraphics[width=1.0\columnwidth, page=1]{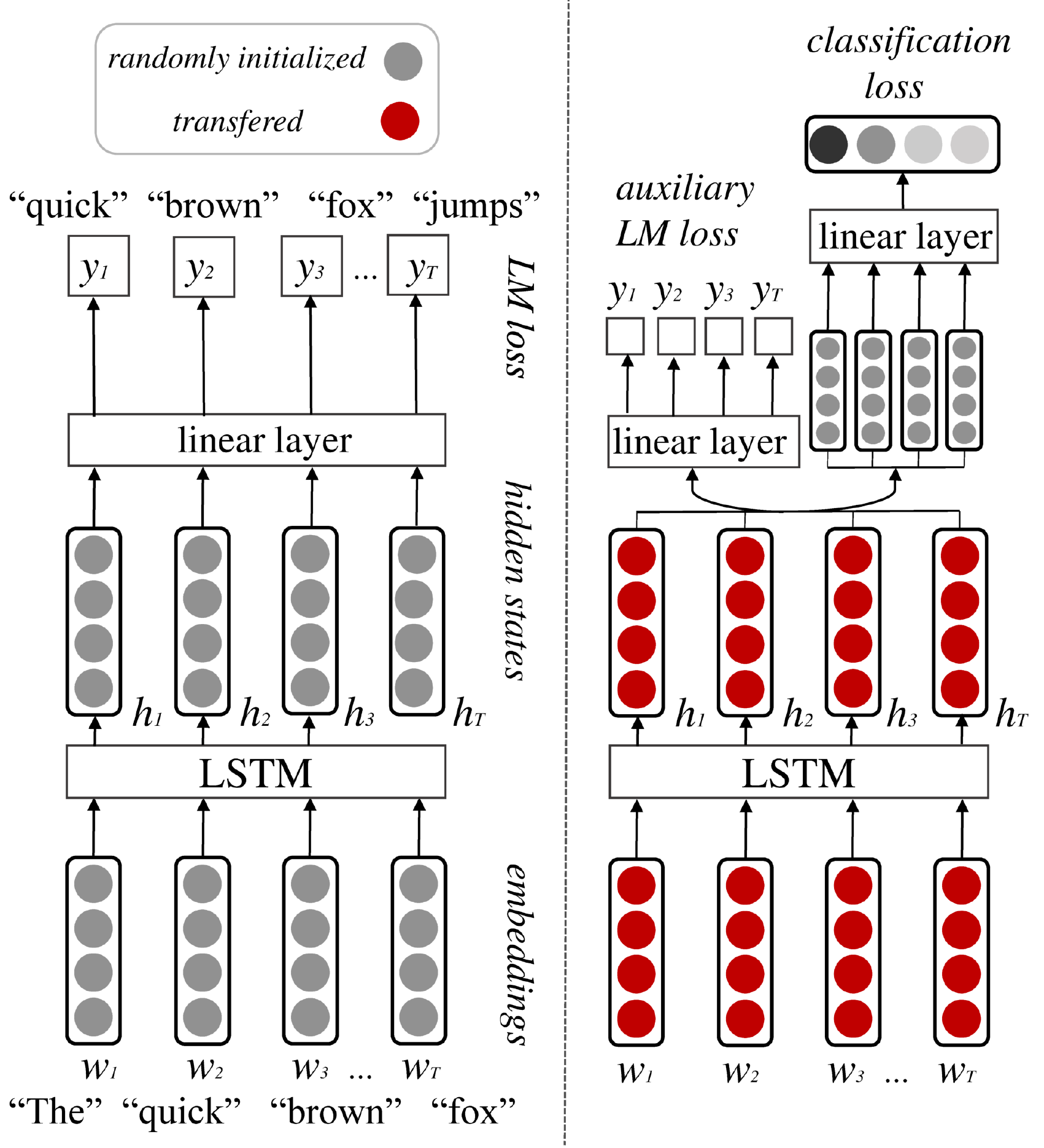}
	\caption{High-level overview of our proposed TL architecture. We transfer the pretrained LM add an extra recurrent layer and an auxiliary LM loss.}
	\label{fig:overview}
\end{figure}
\vspace{5pt}

\noindent In order to adapt the contribution of the pretrained model to the task at hand, we introduce an auxiliary LM loss during training. The joint loss is the weighted sum of the task-specific loss $L_{task}$ and the auxiliary LM loss $L_{LM}$, where $\gamma$ is a weighting parameter to enable adaptation to the target task but at the same time keep the useful knowledge from the source task. Specifically:  
\noindent
\begin{align}
 L &= L_{task} + \gamma L_{LM}  
\end{align}
\noindent

\noindent\textbf{Exponential decay of $\gamma$.}
An advantage of the proposed TL method is that the contribution of the LM can be explicitly controlled in each training epoch.
In the first few epochs, the LM should contribute more to the joint loss of SiATL so that the task-specific layers adapt to the new data distribution. After the knowledge of the pretrained LM is transferred to the new domain, the task-specific component of the loss function is more important and $\gamma$ should become smaller. This is also crucial due to the fact that the new, task-specific LSTM layer is randomly initialized. Therefore, by back-propagating the gradients of this layer to the pretrained LM in the first few epochs, we would add noise to the pretrained representation. To avoid this issue, we choose to initially pay attention to the LM objective and gradually focus on the classification task. In this paper, we use an exponential decay for $\gamma$ over the training epochs. 

\vspace{4pt}

\noindent\textbf{Sequential Unfreezing.}
Instead of fine-tuning all the layers simultaneously, we propose unfreezing them sequentially, according to \citet{howard2018universal,W18-6209}. We first fine-tune only the extra, randomly initialized LSTM and the output layer for $n-1$ epochs. At the $n^{th}$ epoch, we unfreeze the pretrained hidden layers. We let the model fine-tune, until epoch $k-1$. Finally, at epoch $k$, we also unfreeze the embedding layer and let the network train until convergence. The values of $n$ and $k$ are obtained through grid search. 
We find the sequential unfreezing scheme important, as it minimizes the risk of overfitting to small datasets.

\vspace{4pt}

\noindent\textbf{Optimizers.}
While pretraining the LM, we use Stochastic Gradient Descent (SGD). When we transfer the LM and fine-tune on each classification task, we use 2 different optimizers: SGD for the pretrained LM (embedding and hidden layer) with a small learning rate, in order to preserve its contextual information. As for the new, randomly initialized LSTM and classification layers, we employ Adam \cite{kingma2014adam}, in order to allow them to train fast and adapt to the target task.

\begin{table}[h]
\centering
\small\begin{tabular}{lccc} \Xhline{2\arrayrulewidth}
Dataset   &  Domain        & \# classes & \# examples \\ \hline
Irony18          & Tweets        & 4          & 4618        \\
Sent17          & Tweets        & 3          & 61854       \\
SCv2        & Debate Forums  & 2          & 3260        \\
SCv1               & Debate Forums & 2          & 1995        \\
PsychExp          & Experiences   & 7          & 7480   
    \\ \Xhline{2\arrayrulewidth}

\end{tabular}
\caption{Datasets used for the downstream tasks.}
\label{table:datasets}
\end{table}
\vspace{-10pt}
\section{Experiments and Results} 
\subsection{Datasets}

To pretrain the language model, we collect a dataset of 20 million English Twitter messages, including approximately 2M unique tokens. We use the 70K most frequent tokens as vocabulary. We evaluate our model on five datasets: \textit{Sent17} for sentiment analysis \cite{rosenthal2017semeval}, \textit{PsychExp} for emotion recognition \cite{doi:10.1177/053901886025004001}, \textit{Irony18} for irony detection \cite{van2018semeval}, \textit{SCv1} and \textit{SCv2} for sarcasm detection \cite{DBLP:conf/sigdial/OrabyHRHRW16,lukin-walker:2013:LASM}. More details about the datasets can be found in Table \ref{table:datasets}.

\begingroup
\setlength{\tabcolsep}{7pt} 
\renewcommand{\arraystretch}{1.1} 
\begin{table*}[!t]
\centering
\small
\begin{tabular}{l|c|c|c|c|c}
\Xhline{2\arrayrulewidth}
      & Irony18 & Sent17    & SCv2   & SCv1   & PsychExp \\ \hline
BoW                 & 43.7               & 61.0                & 65.1       & 60.9    &    25.8       \\
NBoW      & 45.2                & 63.0                & 61.1       & 51.9     &  20.3         
\\ \hline

P-LM  & 42.7 $\pm$ 0.6   & 61.2 $\pm$ 0.7 & 69.4 $\pm$ 0.4 &  48.5 $\pm$ 1.5 & 38.3 $\pm$ 0.3  \\

P-LM + su  & 41.8 $\pm$ 1.2 & 62.1 $\pm$ 0.8 & 69.9 $\pm$ 1.0 & 48.4 $\pm$ 1.7  & 38.7 $\pm$ 1.0 \\ \hline

P-LM + aux & 45.5 $\pm$ 0.9  & 65.1 $\pm$ 0.6 & 72.6 $\pm$ 0.7 & 55.8 $\pm$ 1.0  & 40.9 $\pm$ 0.5   \\
\textbf{SiATL} (P-LM + aux + su) & \textbf{47.0} $\pm$ 1.1 & \textbf{66.5} $\pm$ 0.2  & \textbf{75.0} $\pm$ 0.7 &\textbf{ 56.8 }$\pm$ 2.0 & \textbf{45.8} $\pm$ 1.6    \\ \hline  \hhline{======}
ULMFiT (Wiki-103)   & $23.6 \pm 1.6$      & $60.5 \pm 0.5$          &\textbf{68.7} $\pm$ 0.6   & \textbf{56.6} $\pm$ 0.5 & 21.8 $\pm$ 0.3     \\
ULMFiT (Twitter)    & \textbf{41.6} $\pm$ 0.7 & \textbf{65.6} $\pm$ 0.4   & 67.2 $\pm$ 0.9  & 44.0 $\pm$ 0.7           &  \textbf{40.2} $\pm$ 1.1    \\ \hline
\multirow{2}{*}{State of the art}  &  53.6    &  68.5  & 76.0         & 69.0       &  57.0\\ \cline{2-6} 
              &   \cite{Baziotis2018NTUASLPAS}          & \cite{cliche2017bb_twtr}   & \cite{ilic2018deep}   &  \multicolumn{2}{c}{\cite{felbo2017using}}       \\ \Xhline{2\arrayrulewidth}
\end{tabular}
\caption{Ablation study on various downstream datasets. Average over five runs with standard deviation. \textit{BoW} stands for Bag of Words, \textit{NBoW} for Neural Bag of Words. \textit{P-LM} stands for a classifier initialized with our pretrained LM, \textit{su} for sequential unfreezing and \textit{aux} for the auxiliary LM loss. In all cases, $F_1$ is employed.}
\label{table:results}
\end{table*}
\endgroup
\subsection{Experimental Setup} \noindent

To preprocess the tweets, we use \textit{Ekphrasis}~\citep{baziotis2017datastories}. For the generic datasets, we use NLTK \cite{bird2004nltk}.
For the \textit{NBoW} baseline, we use \textit{word2vec} \cite{mikolov2013} 300-dimensional embeddings as features. For the neural models, we use an LM with an embedding size of 400, 2 hidden layers, 1000 neurons per layer, embedding dropout 0.1, hidden dropout 0.3 and batch size 32. We add  Gaussian noise of size 0.01 to the embedding layer. A clip norm of 5 is applied, as an extra safety measure against exploding gradients. For each text classification neural network, we add on top of the transferred LM an LSTM layer of size 100 with self-attention and a softmax classification layer. In the pretraining step, SGD with a learning rate of 0.0001 is employed. In the transferred model, SGD with the same learning rate is used for the pretrained layers. However, we use Adam \cite{kingma2014adam} with a learning rate of 0.0005 for the newly added LSTM and classification layers. For developing our models, we use PyTorch \cite{paszke2017automatic} and Scikit-learn \cite{pedregosa2011scikit}.

\section{Results \& Discussion}

\noindent\textbf{Baselines and Comparison.}  Table \ref{table:results} summarizes our results. The top two rows detail the baseline performance of the BoW and NBoW models. We observe that when enough data is available (e.g. \textit{Sent17}), baselines provide decent results. Next, the results for the generic classifier initialized from a pretrained LM (\textit{P-LM}) are shown with and without sequential unfreezing, followed by the results of the proposed model SiATL. SiATL is also directly compared with its close relative ULMFiT (trained on Wiki-103 or Twitter) and the state-of-the-art for each task; ULMFiT also  fine-tunes a LM for classification tasks. The proposed SiATL method consistently outperforms the baselines, the \textit{P-LM} method and  ULMFiT in all datasets. Even though we do not perform any elaborate learning rate scheduling and we limit ourselves to pretraining in Twitter, we obtain higher results in two Twitter datasets and three generic. 

\noindent\textbf{Auxiliary LM objective.}
The effect of the auxiliary objective is highlighted in very small datasets, such as \textit{SCv1}, where it results in an impressive boost in performance (7\%). We hypothesize that when the classifier is simply initialized with the pretrained LM, it overfits quickly, as the target vocabulary is very limited. The auxiliary LM loss, however, permits refined adjustments to the model and fine-grained adaptation to the target task. 

\begin{figure}[!t]
 \centering
 \resizebox{\columnwidth}{!}{%
	\captionsetup{farskip=0pt} 
	\centerline{\includegraphics[width=1\columnwidth]{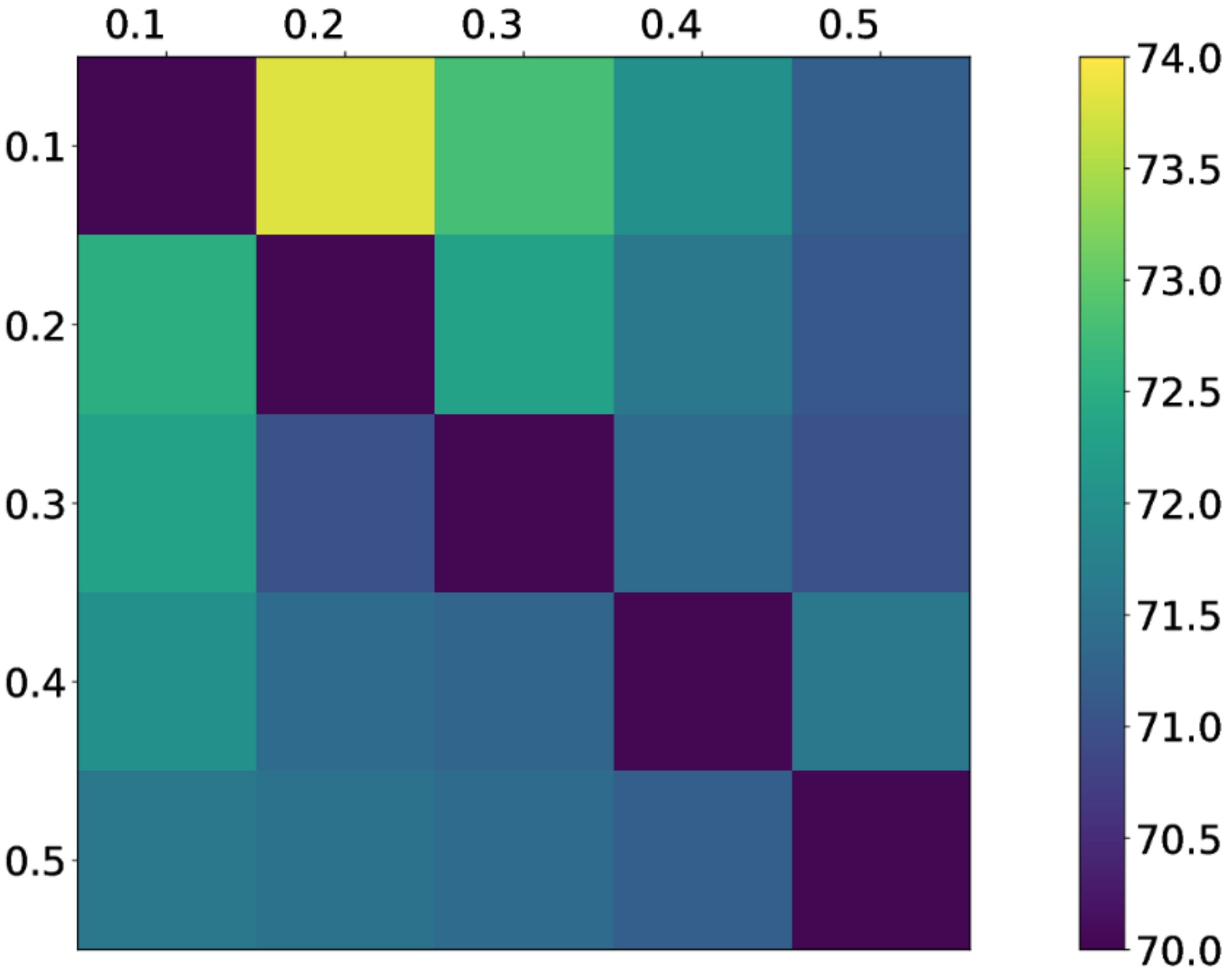}}}
	\caption{Heatmap of the effect of $\gamma$ to $F_1$-score, evaluated on \textit{SCv2}. The horizontal axis depicts the initial value of $\gamma$ and the vertical axis the final value of $\gamma$. }
	\label{fig:heatmap}
\end{figure}

\noindent\textbf{Exponential decay of $\gamma$.}
For the optimal $\gamma$ interval, we empirically find that exponentially decaying $\gamma$ from 0.2 to 0.1 over the number of training epochs provides best results for our classification tasks. A heatmap of $\gamma$ is depicted in Figure \ref{fig:heatmap}. We observe that small values of $\gamma$ should be employed, in order to scale the LM loss in the same order of magnitude as the classification loss over the training period. Nevertheless, the use of exponential decay instead of linear decay does not provide a significant improvement, as our model is not sensitive to the way of decaying hyperparameter $\gamma$.

\noindent\textbf{Sequential Unfreezing.}
Results show that sequential unfreezing is crucial to the proposed method, as it allows the pretrained LM to adapt to the target word distribution. 
The performance improvement is more pronounced when there is a mismatch between the LM and task domains, i.e., the non-Twitter domain tasks. 
Specifically for the \textit{PsychExp} and \textit{SCv2} datasets, sequentially unfreezing  yields significant improvement in $F_1$ building upon our intuition.

 \begin{figure}[!t]
 \centering
 \resizebox{.8\columnwidth}{!}{%
	\captionsetup{farskip=0pt} 
	\centerline{\includegraphics[width=1.3\columnwidth]{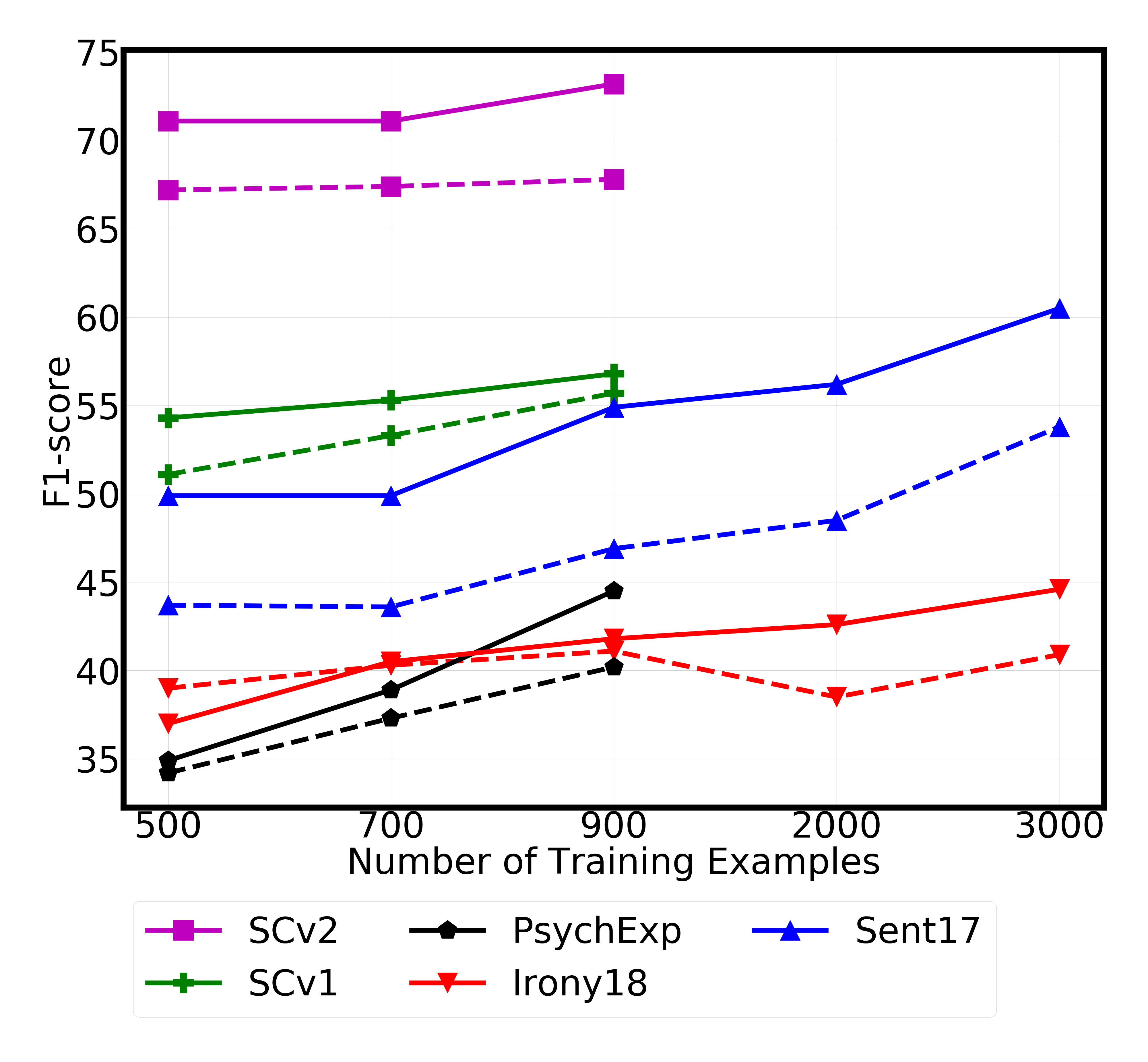}}
	} 
	\caption{Results of SiATL, our proposed approach (continuous lines) and ULMFiT (dashed lines) for different datasets (indicated by different markers) as a function of the number of training examples. }
\label{few}
\end{figure}

\noindent\textbf{Number of training examples.} Transfer learning is particularly useful when limited training data are available. We notice that for our largest dataset \textit{Sent17}, SiATL outperforms ULMFiT only by a small margin when trained on all the training examples available (see Table \ref{table:results}), while for the small \textit{SCv2} dataset, SiATL outperforms ULMFiT by a large margin and ranks very close to the state-of-the-art model \cite{ilic2018deep}. Moreover, the performance of SiATL vs ULMFiT as a function of the training dataset size is  shown in Figure~\ref{few}. Note that the proposed model achieves competitive results on less than 1000 training examples for the \textit{Irony18, SCv2, SCv1} and \textit{PsychExp} datasets, demonstrating the robustness of SiATL even when trained on a handful of training examples.

\noindent\textbf{Catastrophic forgetting.} We observe that SiATL indeed provides a way of mitigating catastrophic forgetting. Empirical results that are shown in Table \ref{table:results} indicate that by only adding the auxiliary language modeling objective, we obtain better results on all downstream tasks. Specifically, a comparison of the  \textit{P-LM + aux} model and the \textit{P-LM} model shows that the performance of SiATL on classification tasks is improved by the auxiliary objective. We hypothesize that the language model objective acts as a regularizer that prevents the loss of the most generalizable features.
 
\section{Conclusions and Future Work}

We introduce SiATL, a simple and efficient transfer learning method for text classification tasks. Our approach is based on pretraining a LM and  transferring its weights to a classifier with a task-specific layer. The model is trained using a task-specific functional with an auxiliary LM loss. 
SiATL avoids catastrophic forgetting of the language distribution learned by the pretrained LM. Experiments on various text classification tasks
yield competitive results, demonstrating the efficacy of our approach. Furthermore, our method outperforms more sophisticated transfer learning approaches, such as ULMFiT in all tasks.

In future work, we plan to move from Twitter to more generic domains and evaluate our approach to more tasks. 
Additionally, we aim at exploring ways for scaling our approach to larger vocabulary sizes~\cite{kumar2018von} and for better handling of out-of-vocabulary words (OOV) ~\cite{DBLP:journals/corr/abs-1804-08205, sennrich2015neural} in order to be applicable to diverse datasets.

Finally, we want to explore approaches for improving the adaptive layer unfreezing process and the contribution of the language model objective (value of $\gamma$) to the target task. 

\section*{Acknowledgments}

We would like to thank Katerina Margatina and Georgios Paraskevopoulos for their helpful suggestions and comments. This work has been partially supported by  computational time granted from the Greek Research \& Technology Network  (GR-NET) in the National HPC facility - ARIS. Also, the authors
would like to thank NVIDIA for supporting this
work by donating a TitanX GPU.

\bibliography{naaclhlt2019}
\bibliographystyle{acl_natbib}
\appendix

\end{document}